\documentclass{isprs}
\usepackage{subfigure}
\usepackage{setspace}
\usepackage{geometry} 
\usepackage{epstopdf}
\usepackage{amsmath}
\usepackage{natbib}
\usepackage{amssymb}
\usepackage[labelsep=period]{caption}  
\usepackage{tikz}
\usepackage{xcolor}
\usepackage{color}
\usepackage[labelsep=period]{caption}  
\usepackage{soul}
\usepackage{multicol}
\usepackage{float}
\usepackage{booktabs}
\usepackage{diagbox}
\usepackage{enumitem}
\usepackage{bm}
\usepackage{paralist}

\begin{document}

\title{INFERRING ROUTING PREFERENCES OF BICYCLISTS FROM SPARSE SETS OF TRAJECTORIES}

\author{
 J. Oehrlein\textsuperscript{1}, A. F\"orster\textsuperscript{1}, D. Schunck\textsuperscript{1}, Y. Dehbi\textsuperscript{1,}\thanks{Corresponding author}\hspace{2mm}, R. Roscher\textsuperscript{2}, J.-H. Haunert\textsuperscript{1}}


\address{
	Institute of Geodesy and Geoinformation, University of Bonn\\ \textsuperscript{1 }(oehrlein, foerster, schunck, dehbi, haunert)@igg.uni-bonn.de, 
  \textsuperscript{2 }ribana.roscher@uni-bonn.de}



\icwg{}   


\abstract{Understanding the criteria that bicyclists apply when they choose their routes is crucial for planning new bicycle paths or recommending routes to bicyclists. 
This is becoming more and more important as city councils are becoming increasingly aware of
limitations of the transport infrastructure and problems related to automobile traffic.
Since different groups of cyclists have different preferences, however, 
searching for a single set of criteria is prone to failure.
Therefore, in this paper, we present a new approach to classify trajectories recorded and shared by bicyclists into different groups
and, for each group, to identify favored and unfavored road types.
Based on these results we show how to assign weights to the edges of a graph representing the road network
such that minimum-weight paths in the graph, which can be computed with standard shortest-path algorithms, correspond to adequate routes.
Our method combines known algorithms for machine learning and the analysis of trajectories in an innovative way 
and, thereby, constitutes a new comprehensive solution for the problem of deriving routing preferences from initially unclassified trajectories.
An important property of our method is that it yields reasonable results even if the given set of trajectories is sparse in the sense that it does not cover all segments of the cycle network.}


\maketitle


\section{INTRODUCTION}\label{intro}

\begin{figure}[h]
	\centering
		\includegraphics[width=0.7\columnwidth]{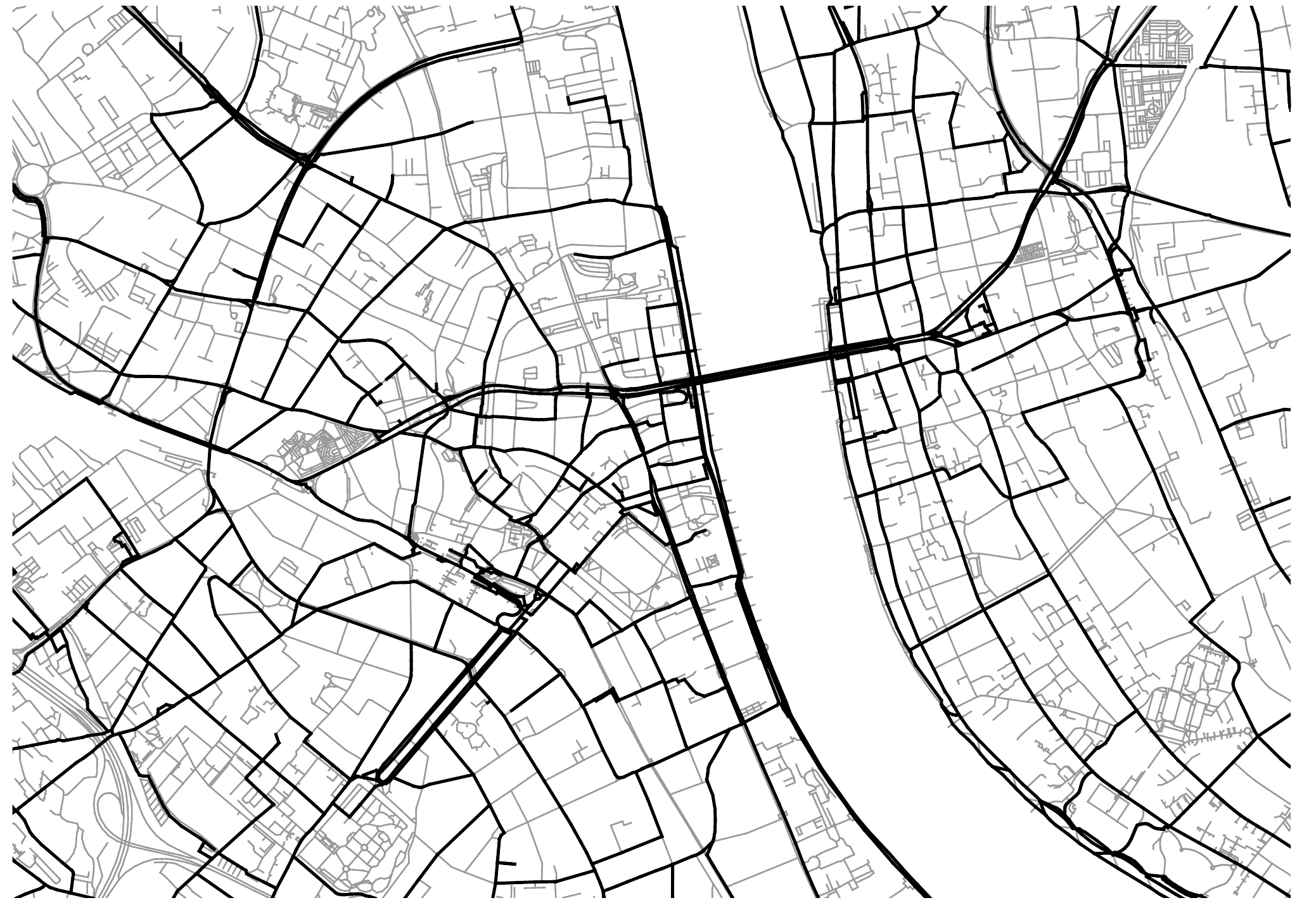}
	\caption{The road segments for a small part of our test area, classified into segments that were used by at least one trajectory (black) and those that were not used (gray). The latter includes 75.81\% of all edges and 68.70\% of their total length.}
	\label{fig:extract}
\end{figure}

Faced with the problem of organizing the traffic in rapidly growing cities, 
many city planners try to support cycling as an environmentally friendly and healthy means of 
transport.
In order to increase the attractiveness of cycling, 
methods for analyzing the routes that bicyclists prefer are needed.
Spatial information that is collected by volunteers (i.e., volunteered geographic information) 
can be used to establish a rich data basis for such methods.
In particular, trajectories that cyclists record and share via on-line platforms (e.g., GPS tracks from Strava or GPSies)
provide information that is not available from other sources. 
Extracting the information about routing preferences in a meaningful form
is far from trivial, however, since the data sets have to be subdivided 
(e.g., to analyze routing preferences separately for different groups of cyclists)
or integrated (e.g., to enrich GPS tracks with information on road types).
Therefore, a methodology that combines multiple data sources and algorithms is needed.
A problem that has not been sufficiently addressed yet is how routing preferences can be inferred
if the given set of trajectories is \emph{sparse} in the sense that the 
trajectories do not cover all segments of the cycle network -- see the extract from the input data that we used in our experiments in Fig.~\ref{fig:extract}.
Still, it is a desirable goal to learn the routing preferences of an individual or a group of cyclists in a form
that allows the computation of an optimal path between any two locations in the network.
In this paper, we present a new methodology to achieve this goal.

Our methodology for inferring routing preferences of cyclists from multiple sources requires trajectories (e.g., GPS tracks), land-use information, and a road network model in the form of a graph $G = (V,E)$ as input -- the latter should include relevant bicycle paths as well as information on road types, such that roads that are forbidden for cyclists (e.g., motorways) can be removed and influences of road types on route choices can be analyzed.
The ultimate goal is to determine an edge weighting $w\colon E \rightarrow \mathbb{R}_{\ge 0}$ for each individual cyclist, to reflect his or her personal routing preferences, or at least one edge weighting for each group of cyclists (e.g., mountain bikers, racing cyclists, and others) that we can identify with the available data.
Generally, in the context of this paper, the weight $w(e)$ of an edge $e \in E$ is interpreted as the cost for traversing $e$ -- the edge weights are equal to the lengths of the edges if the cyclists simply prefer short routes, but other weight settings 
are needed to express that cyclists accept detours in order to avoid unfavorable road segments (e.g., unpaved trails in the case of racing cyclists). 
We also write $w(P)$ to refer to the total weight of a path $P$.
Since a weighted graph model is required as input by most routing algorithms, 
the outcome of our method can be used to infer user-dependent or group-dependent optimal paths between any two locations in the cycle network.
This could be useful for cyclists who use bicycle navigation systems for route planning, 
but also for spatial planners who conduct shortest-path analyses with geo-information systems, e.g., to predict traffic loads for planned 
bicycle paths.


\begin{figure}
	\centering
		\includegraphics[width=\columnwidth]{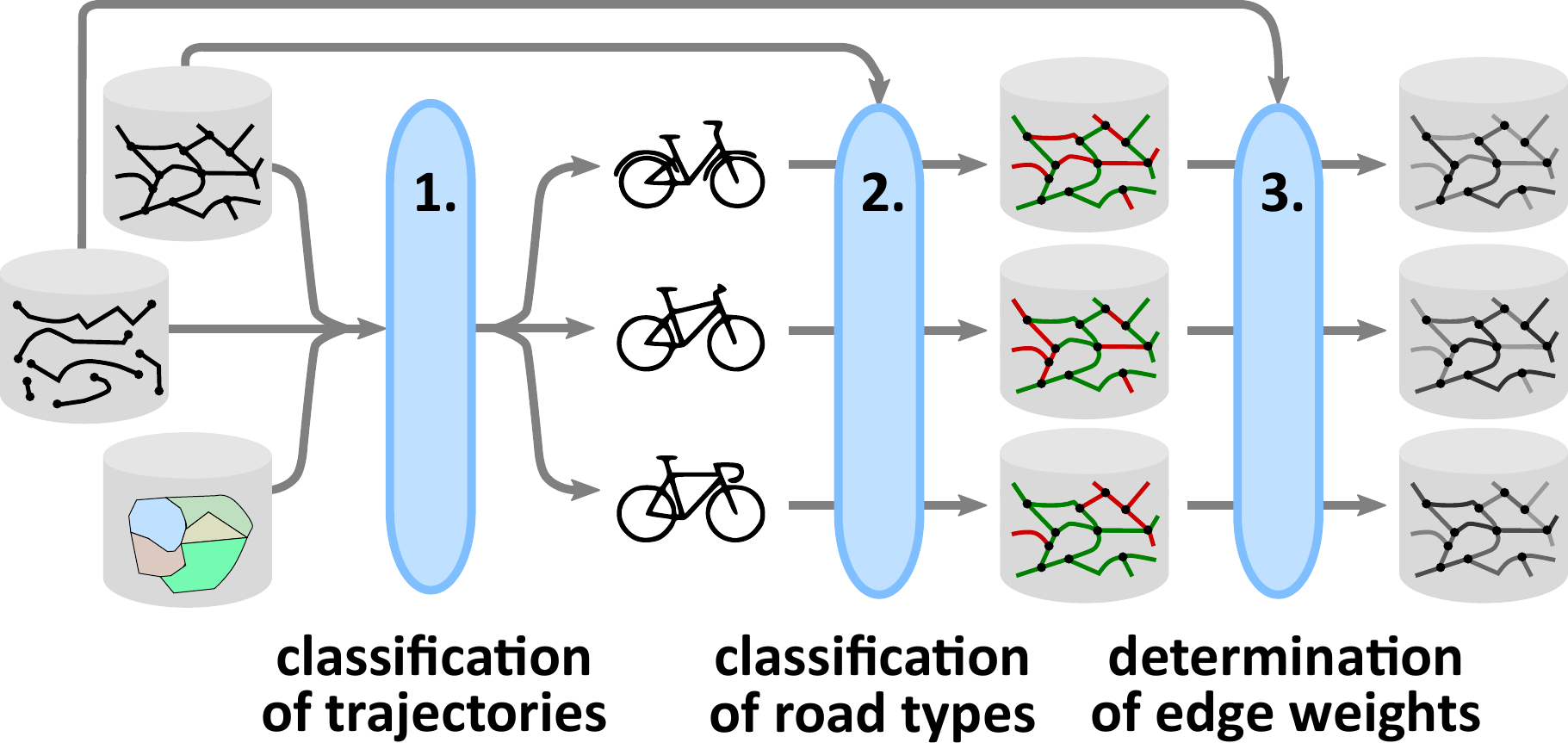}
	\caption{An overview of our method for the identification of cyclist groups and their routing preferences. After a data acquisition from different sources, a multi-source data analysis consisting of three steps is performed. }
	\label{fig:overview}
\end{figure}


Three steps are conducted in order to get from the source data to the weighted graph models.
These steps are illustrated in Fig.~\ref{fig:overview} and specified below.

\begin{compactenum}
\item In the first step, the different information sources are combined using a map-matching algorithm. Geometric buffering operations as well as shortest-path computations (with a default weight setting) are applied to extract 
meaningful features for an unsupervised classification method.
This is used to classify the trajectories with respect to different cyclist groups.
\item In the second step, favored and unfavored road types are identified for each group. In this context, a high proportion of a certain road type within the trajectories of a group is an indicator of preference of this road type. 
\item In the third step, we learn a function that maps edge types 
and edge lengths to edge weights.
Note that this function yields a weight for \emph{every} edge, no matter whether or not it is used by any of the trajectories.
The edge weights can be used to consider the routing preferences when computing new routes.
We will show how to conduct this step for each single trajectory as well as for the set of trajectories of each cyclist group.
\end{compactenum}

The main contribution of this paper is the automatic identification of a method for identifying favorable and unfavorable road types
and the computation of a weighted graph model for a group of users that reflects the 
preferences of the road types.

The remainder of this paper is structured as follows. 
We review related work in Sect.~\ref{sec:review} and 
present our methodology in detail in Sect.~\ref{sec:method}.
Then, we discuss our experimental results in Sect.~\ref{sec:experiments} and conclude the paper in Sect.~\ref{sec:conclusion}.

\section{RELATED WORK}\label{sec:review}

Analyzing trajectories has become a major research discipline within computer science and geographic information science --
we do not aim to give a comprehensive overview but refer to the survey article by \cite{mazimpaka2016trajectory}.
An important task of trajectory analysis is to improve navigation systems and routing algorithms based on trajectories recorded by users.
Based on taxi trajectories, \cite{Yuan2010} were able to infer travel times for road segments and, thus, 
to enable the computation of fastest routes for cars.
More generally, trajectories can be used to augment a network model with additional attributes acquired by users, 
for example, with information on road surface quality inferred from accelerometer data of bicyclists \citep{Reddy2010}.
\cite{Kessler2013} and \cite{SultanBHD17} have discussed in detail how volunteered geographic information (VGI) can be used to analyze bicycle routes.
An open problem is still, however, to learn previously unknown routing criteria and to adopt them for 
the application in routing algorithms.

In the context of understanding cyclists' behavior and their routing preferences, \cite{broach2012cyclists} observed 164 cyclists over a couple of days. They analyzed their practices based on recorded GPS trajectories. The distance, turn frequency, slope, intersection control and traffic volumes turn out to be the major parameters influencing the choice the cyclist trip paths. Infrastructures such as off-street bike paths as well as the trip category, e.g. commute or utilitarian, have also an impact on the route preferences of the investigated tracks. In contrast to \cite{broach2012cyclists}, 
we are particularly interested in analyzing how route choices are influenced by road types.
Furthermore, we aim at the automatic identification of cyclist groups from crowd-sourced data and the learning of a routing model for each detected group based on its specific preferences. 

In order to derive a weighted graph model reflecting the routing preferences of cyclists, \cite{Bergmann2016} have chosen a rather pragmatic approach by counting for each road segment the 
number of users and the number of trajectories using it.
Based on these numbers, they have defined three different measures to derive edge weights, 
all of which are based on the assumption that highly used road segments should receive low weights.
We argue, however, that the frequency of usage should not directly 
be translated into an edge weight. Consider for example a triangular graph whose edges represent connections between three cities $A$, $B$, and $C$.
If we observe a large amount of traffic on the edge $\{A,B\}$ connecting $A$ and $B$, we must not conclude 
that this edge is by any means `cheaper' than $\{B,C\}$ or $\{C,A\}$ and that it should receive a low weight. 
Instead, the high usage of edge $\{A,B\}$ might also be due to the fact that many people commute between $A$ and $B$.
Furthermore, inferring the weights of edges from the frequency of their usage fails if the set of trajectories is sparse.

Probably the first method that infers routing preferences 
from a sparse set of trajectories has been presented by 
\cite{Balteanu2013}.
As all methods that we review in the following, it even works if only a single trajectory is
provided as input. 
More precisely, given a graph $G$ with two edge weightings $w_1$, $w_2$ (e.g., travel time and geometric distance) and a user's path $P$ (the trajectory) between two nodes $s$ and $t$ in $G$,
the aim is to infer a parameter $\beta$ 
such that an $s$-$t$-path $P'$ minimizing 
\begin{equation}\max\{ \beta \cdot w_1(P'), (1-\beta) \cdot w_2(P')\}
\label{eq:pareto}
\end{equation} 
is most similar to $P$. 
Without reviewing in detail how similarity is defined in this context, 
we note that the parameter $\beta$ inferred by the method can indeed explain how the user trades off between $w_1$ and $w_2$.
A clear disadvantage of the method of \cite{Balteanu2013} with respect to its practical relevance
is, however, that the trained routing model is of little use if the aim
is to compute new routes with standard routing algorithms (e.g.,
the algorithm of \citet{Dijkstra1959})
which, usually, require a single edge weighting as input.
More precisely, a standard routing algorithm does not yield a path $P'$ 
minimizing (\ref{eq:pareto}) for a trained parameter $\beta$.
Therefore, in the following, we focus on methods that 
try to define a new edge weighting $w$ based on a linear combination of the given edge weightings.
After the coefficients of the linear combination have been learned and, thus,
the new edge weighting is fixed, one can use standard routing algorithms to compute routes that
are optimal with respect to $w$.

\cite{FunkeEtAl2016} have studied the problem in which a graph $G$ with multiple edge weightings $w_1, \dots, w_d$  as well as a path $P$ between two nodes $s$ and $t$ in $G$ 
are given as input and the aim is to compute a linear combination $w = \alpha_1 \cdot w_1+ \dots+ \alpha_d \cdot  w_d$ of the weightings
such that $P$ is a weight-minimal $s$-$t$-path with respect to the new weighting $w$.
This weighting is assumed to represent the routing preferences of a user who chose $P$ as his or her route.
Unfortunately, the problem can be infeasible for a path corresponding to the trajectory of a user,
since the path may not be optimal with respect to any weighting.
Funke et al.\ address this issue by suggesting that if the problem is infeasible for a given path
then the path should be divided into two subpaths of equal length and the problem should be solved independently for
each of the two subpaths (which many require further recursive splitting to end up with feasible problem instances). With this approach, however, artificial split points are introduced and different linear combinations are obtained 
for the different subpaths.

The algorithm of \citet{OehrleinNH17} is similar to the one of \cite{FunkeEtAl2016} in the sense
that it computes a partition of a given path into multiple subpaths and a linear combination of different weightings. However, the partition and the new weighting are computed such that \emph{all} of the resulting subpaths are optimal with respect to the \emph{same} weighting $w$, meaning that the different subpaths are not considered as independent 
problem instances. Furthermore, instead of partitioning a path into two subpaths of equal lengths,
the algorithm of \citet{OehrleinNH17} uses an optimization criterion to decide where to introduce split points.
More precisely, given a graph $G$ with two edge weightings $w_1$ and $w_2$ and a 
path $P$ in $G$, the algorithm yields a new weighting $w = \alpha \cdot w_1 + (1-\alpha) \cdot w_2$
and a partition of $P$ into a \emph{minimum number} of subpaths such that each of the subpaths is
optimal according to the weighting $w$.
Compared to the algorithm of \cite{FunkeEtAl2016}, the algorithm of \citet{OehrleinNH17}
is certainly more advanced with respect to how it computes the split points, but it
has the disadvantage that it can deal with only two given weightings $w_1$ and $w_2$
and not with an arbitrary number $d$ of weightings.
Nevertheless, we choose this method since inferring a trade off between two criteria from
a sparse set of trajectories is already challenging.
In the following, we refer to the split points computed by the method
as \emph{milestones} and the partition of the given path induced by the split points
as a \emph{milestone decomposition}.

An important property of the method of \citet{OehrleinNH17}
is that it not only computes the parameter $\alpha$
corresponding to an optimal milestone decomposition
but that it systematically explores different values for $\alpha$
and tests the effect on the size of the milestone decomposition.
This offers new possibilities of studying the quality of a bi-criteria routing model
as a function of its trade-off parameter $\alpha$, which we will show in Sect.~\ref{sec:experiments} for 
the experiments that we conducted.

\section{Methodology}\label{sec:method}

In this section we present the mathematical foundations of our method,
including the routing model whose parameter we aim to learn (Sect.~\ref{sec:model})
and the concepts behind each of the three steps of our method (Sections~\ref{sec:classification}--\ref{sec:edgeweights}).

\subsection{Routing Model} \label{sec:model}

A meaningful representation of a user's routing preferences in a given graph 
$G = (V, E)$ is a weighting $w \colon E \rightarrow \mathbb{R}_{\ge 0}$ that assigns to
each edge $e \in E$ a weight $w(e)$. This can be assumed to represent a cost for traversing edge $e$.
Our aim is to learn such a weighting from trajectories,
which will allow us to compute optimal paths for a user or group of users with known algorithms, for example, with
the classical algorithm of \citet{Dijkstra1959} or with the help of modern speed-up techniques, such as contraction hierarchies \citep{GeisbergerSSD08}.

Since the given graph $G$ may not be completely covered by the trajectories, 
the trajectories alone do not suffice to infer weights for all edges of $G$.
Therefore, a good strategy is to define the weighting based on attributes that are given for each edge 
(for example, its length and road type) and to use the trajectories only to infer 
a small number of parameters that condense the attributes into a weight.
In this paper, we suggest a model that requires for each user group a classification of road types 
into unfavorable and favorable types (for example, arterial streets and bicycle paths, respectively)
and, additionally, a single parameter $\alpha \in [0,1]$.
According to this model, the weight of edge $e \in E$ is
\begin{equation}
w(e) = \left\{\begin{array}{lr} \alpha \cdot \mathrm{length}(e) & \mathrm{if} \quad e \in E^+ \\ (1 - \alpha) \cdot \mathrm{length}(e) & \mathrm{if} \quad e \in E^- \end{array}\right. \label{eq:weight}
\end{equation}
where $\mathrm{length}(e)$ is the length of $e$ and $\{E^-, E^+\}$ is a binary classification of $E$
into a set $E^-$ of edges with an unfavorable road type and a set $E^+$ of edges with a favorable road type.
This model implies that traversing an edge with an unfavorable type is by factor $\frac{1-\alpha}{\alpha}$ 
more expensive than traversing an edge of the same length with a favorable type.
Obviously, one would expect $\alpha \le 0.5$, since otherwise edges with unfavorable types would be preferred,
which would be a contradiction.
However, we leave it to the inference algorithm that we apply (see Sect.~\ref{sec:edgeweights})  to select $\alpha \in [0,1]$
and, afterward, test for $\alpha \le 0.5$ to check the consistency of the result.
To summarize, we need to detect different groups of users and for each group the binary classification $\{E^-, E^+\}$ as well as the parameter $\alpha$.

Obviously, this approach could be generalized by classifying the road types into more 
than two classes. The more classes are considered, however, 
the more parameters would have to be learned in order to derive the edge weights from the types and the geometric lengths of the edges.
Since many road segments are not covered by any trajectory and since for some road types only few road segments exist, 
inferring a binary classification and learning the parameter $\alpha$ for each user group is already challenging. 
Nevertheless, learning a more sophisticated model is an interesting task for future research.
Since the algorithm of \citet{OehrleinNH17} 
that we apply in our workflow is currently limited to two edge weightings,
however, such an improvement would require a more substantial innovation.

\subsection{Classification of Trajectories}\label{sec:classification}

In the first step of our method, a multi-source data analysis is performed in order to automatically classify the set of trajectories with respect to different cyclist groups. To this aim, openly accessible GPS-tracks are collected from a user-driven platform. The trajectories are then augmented by additional information such as road types. The extraction of additional features is performed after a map-matching process, which establishes correspondences between a given trajectory and an underlying road network of the region of interest. In order to achieve an accurate analysis, the trajectories are also enriched by information about the surrounding areas stemming from a digital landscape model. This, in particular, gives insight into the land-use categories of the areas through which the trajectories pass.
Furthermore, for each trajectory, a path of minimum length is computed connecting the trajectory's source and destination.
This yields additional interesting features, such as the length ratio between the trajectory and the optimal path (also known as the detour factor or dilation).
All this information is serving as a rich feature set for the extraction of cyclist groups in an unsupervised learning process.


The classification of the trajectories into meaningful cyclist groups is done in an unsupervised way using
the $k$-means algorithm \citep{lloyd1982least}.
Although the users normally specify the types of their trajectories when they share them and, 
thus, a user-specified grouping of the trajectories is available, the assignment is subjective and partly erroneous, which is underlined by our experiments (see Sec.~\ref{exp:classification}).
Therefore, in the subsequent steps of our method, we use the result of the unsupervised classification algorithm instead of the user-specified types.

\subsection{Recognizing Unfavorable and Favorable Road Types}

Road networks are usually represented as sets of line segments with associated road types.
Therefore, as a result of the map-matching process, we obtain for each user group the distribution of road types over the total length of the trajectories.
Although such statistics give interesting insights, we have to be careful
of what we conclude.
Suppose that the type ``residential street'' constitutes 95\% of the total length of
a group's trajectories.
This high percentage may either be due to the fact that the group considers residential streets as favorable or because there is a lack of bicycle-friendly paths and, thus, the users had to choose an unfavorable road type. 
Therefore, we compute for each trajectory the geometrically shortest path in the road network connecting the trajectory's start and end point and use that path as a reference.
For each road type $c$, we compare the relative share $r_\mathrm{user}(c)$ of $c$ among the total length of the trajectories with the relative share $r_\mathrm{shortest}(c)$ of $c$ among the total length of the shortest paths.
If $r_\mathrm{user}(c) > r_\mathrm{shortest}(c)$, one may argue that the users had the possibility of using 
shorter paths but decided to use longer paths with a larger share of type $c$. 
This can be seen as an indication of $c$ being a favorable type. Consequently, we define
\begin{equation}
E^+ = \{e \in E \mid r_\mathrm{user}(c(e)) \ge r_\mathrm{shortest}(c(e)) \} \label{eq:favored}
\end{equation}
\begin{equation}
E^- = \{e \in E \mid r_\mathrm{user}(c(e)) < r_\mathrm{shortest}(c(e)) \}\,, \label{eq:unfavored}
\end{equation}
where $c(e)$ is the road type of edge $e$.

\subsection{Inferring Edge Weights} \label{sec:edgeweights}

Generally, an edge weighting $w$ alone can not explain the trajectories of a user group
since, for example, even within one group different criteria are applied or the trajectories include round trips 
that were clearly not chosen as minimum-cost paths between two nodes.
Moreover, the model that we introduced with Equation~(\ref{eq:weight}) may be too restrictive to subsume the
weighting actually applied by a user.
Nevertheless, we aim to determine the parameter $\alpha$ such that the model explains the trajectories of a user group \emph{as much as possible}.
For this purpose, we apply the algorithm by \cite{OehrleinNH17}.
Recall that, given a user's trajectory $T$ as a path in a graph $G = (V, E)$ with two edge weightings $w_1$ and $w_2$, this algorithm partitions $T$ into a minimum number of sub-trajectories and, simultaneously, selects a parameter $\alpha \in [0,1]$, 
such that each of the resulting sub-trajectories is an optimal path in $G$, 
in the sense that no path connecting the same two nodes is better according to the weighting $w = \alpha \cdot w_1 + (1-\alpha) \cdot w_2$.
Since minimizing the number of sub-trajectories is the same as maximizing their average length, 
the weighting $w$ that is learned with the method explains the routes chosen by the users relatively well.

\citet{OehrleinNH17} used their algorithm to understand how slope affects the route choice of bicyclists.
With our model, however, where the weighting $w$ should reflect unfavorable and favorable road types, the algorithm
needs to be applied with the following setting:
\begin{equation}
w_1(e) = \left\{\begin{array}{lr} \mathrm{length}(e) & \mathrm{if} \quad e \in E^+ \\ 0 & \mathrm{if} \quad e \in E^- \end{array}\right. \label{eq:weight1}
\end{equation}
\begin{equation}
w_2(e) = \left\{\begin{array}{lr} 0 & \mathrm{if} \quad e \in E^+ \\  \mathrm{length}(e) & \mathrm{if} \quad e \in E^- \end{array}\right. \label{eq:weight2}
\end{equation}
With this setting, $\alpha \cdot w_1 + (1-\alpha) \cdot w_2$ is indeed equal to $w$ as defined in Equation~(\ref{eq:weight}).

The algorithm of \citet{OehrleinNH17} requires integer weights as input, which we ensure by rounding the edge lengths to m.
It works by systematically testing different values $\alpha \in [0,1]$, 
including the interval boundaries 0 and 1. We encountered very long running times for those extremal values and, 
therefore, decided to restrict the search to $\alpha \in [0.1,0.9]$. 
With this we still take into consideration that, in order to avoid an unfavorable edge $e \in E^-$, a user may accept a detour of nine times the length of $e$.
However, longer detours are not considered.

\section{EXPERIMENTS}\label{sec:experiments}

This section presents our conducted experiments and experimental results and gives insight into the data used in the different steps of our approach.

\subsection{Data}
Since Bonn is representing an example of a bicycle-friendly city, we decided to demonstrate our approach for this region. 
82\% of the households in Bonn are owning at least one bicycle. Furthermore, not only the city but also the surrounding areas, for instance ``Siebengebirge'' and the bank of the Rhine river, are attractive for bicycle tours. 
The Bonn's city council is striving till 2020 to declare Bonn as the capital of bicyclists in the federal land of North Rhine Westphalia. 

Our experiments are performed on crowd-sourced data stemming from the user-driven platform GPSies \footnote{GPSies.com}. 
From this platform, GPS trajectories, which have been recorded by users with different preferred activities, can be downloaded. 
In our context, we are especially interested in the following three types of cyclist activities: \texttt{biking}, \texttt{mountainbiking}, and \texttt{racingbiking}. 
For the evaluation of our algorithm, we downloaded about 250 trajectories for each user group from the region of Bonn and surroundings in Germany. 
Beside the GPS-coordinates, each track contains additional information about the whole length, climb and descent of the trajectory. 
Furthermore, two types of tracks are discriminated: circular and simple tracks. We denote these features as \texttt{feature set a}.

In order to analyze these trajectories, we extracted additional features from road segments corresponding to the underlying trajectories. The correspondences were computed with the map matching algorithm of \citet{HaunertBudig2012}. To this aim, we used a road network of the same area from \texttt{OpenStreetMap}\footnote{OpenStreetMap.org} (\texttt{OSM}). 
We denote these features as \texttt{feature set b}.

In this way, each trajectory path is augmented by the information acquired from the associated road segment from \texttt{OSM}. 
For our purpose, the street type category including among others roads, paths and cycle tracks is of great relevance.
The inferred information enables for example a trajectory analysis depending on the used street types for each cyclist group.  

In order to learn different weightings for different user groups of cyclists, additional information about a given trajectory and its surrounding is needed. 
Thus, we exploited data related to our region of interest stemming from the German Digital Landscape Model ATKIS-DLM \footnote{www.opengeodata.nrw.de}.
The latter is an object-based vector model which defines an object set with several object types accordingly. 
The object types comprise for instance \textit{woodland}, \textit{arable land} and \textit{settled land}. 

\subsection{Results of the Trajectory Classification}\label{exp:classification}
In this experiment, we classify the trajectories into specific activity groups and compare them to the user-provided groups.
For this, we use the provided information from both\texttt{ feature set a }as well as\texttt{ feature set b}.
The features are z-normalized to zero mean and unit standard deviation to ensure an equal weighting of each single feature. 

We cluster the data utilizing k-means and manually assign activity groups to the resulting clusters. 
We run k-means with different initializations and choose the result with the highest compactness. We choose different numbers of clusters, and manually decide on the best number by means of the quality of the assignment. 

Furthermore, we determined the importance of specific features using the reliefF algorithm \citep{Kononenko1994} and analyzed the influence of restricting the set of features to the most important ones on the k-means clustering result.  
We manually tested several values for the amount of neighbors necessary to calculate the importance for each feature, and report the results for $k_{\text{reliefF}}=100$. 
We observed that for larger values the set of the most important features converges to a fixed set. 
We used the calculated weights and determined all features that lie within the $90\%$-quantile.

The evaluation of different numbers of clusters confirm the user-provided groups such that k-means 
provide the best interpretable result using three clusters with features which can be assigned to the user-provided groups.
Table~\ref{tab:conf_all} shows the contingency table, which is a detailed analysis of the number of trajectories assigned to the three different groups \texttt{racingbiking}, \texttt{mountainbiking}, and \texttt{biking} by the user and by k-means.
The table also includes information about the number of trajectories assigned to the same group and assigned to different groups by the user and by k-means.

\begin{table}[ht]
\resizebox{\linewidth}{!}{%
\begin{tabular}{lccccc}
\toprule
\backslashbox{clustered}{original} & \texttt{mountainbiking} & \texttt{racingbiking} & \texttt{biking} & sum
\\ \midrule \midrule
\texttt{mountainbiking}      & 125            & 20         & 39     & 184 (68\%) \\
\texttt{racingbiking}          & 10              & 135        & 47     & 192 (70\%) \\
\texttt{biking}              & 17             & 63         & 141    & 221 (64\%) \\
sum               & 152 (82\%)            & 218 (62\%)        & 227 (62\%)   & 597 (67\%) \\ \bottomrule
\end{tabular}%
}
\caption{Contingency table, showing the number of trajectories assigned by the user and by k-means clustering to \texttt{biking}, \texttt{mountainbiking} and \texttt{racingbiking}.}
\label{tab:conf_all}
\end{table}

Overall, there is a consensus in 67\% of all trajectories.
There is an increase for \texttt{mountainbiking} and a decrease for \texttt{racingbiking} after k-means is applied.
Although nearly the same numbers of trajectories are assigned to \texttt{biking}, this group shows the largest difference in our comparison. 
Around 40\% of users which assign themselves to the group \texttt{biking} are classified as a different group by k-means.
Especially users who classify themselves as part of the group \texttt{racingbiking} are assigned to \texttt{biking} by k-means.
For a more detailed examination, we choose different trajectories which have a different assignment by the user and by k-means, and analyzed them by means of various features. 
It turned out that in most cases users assign themselves to an activity group which does not fit their biking behavior, or the trajectory's features lie close to the cluster boundary.

Finally, we analyzed the features' importance obtained by reliefF. 
The sorted list of the most important features in decreasing order is (1) the route type (circular or simple track), (2) the altitude range, (3) the difference between the length of the actual trajectory and the shortest path-trajectory, (4) percentage of agricultural area close the trajectory, (5) percentage of forest close the trajectory, followed by multiple features defining the road type, and the living environment.
We repeated k-means clustering with the most important features and compared it to the clustering results using all features.
Both clustering results agree in $96\%$ of all trajectories.
Moreover, we compared the contingency table obtained by k-means with all features (cmp. Tab. \ref{tab:conf_all}) and the contingency table obtained by k-means with the most important features, and receive a mean absolute difference of 3.35\%.
Both result indicate that the identified most important features describe the activity groups well.

\subsection{Results of the Road-Type Classification}

In this experiment we compute for each trajectory $T$ a shortest path $P$ in the road network
that connects the start node and end node of $T$.
For each group of bicyclists, we analyze the share of the different roads types among the total length of all trajectories as well as among the total length of all shortest paths.
A comparison allows us to infer which of the road types are favored and unfavored.

\begin{figure}
\includegraphics[trim={0 0 4px 0}, clip, width=\columnwidth]{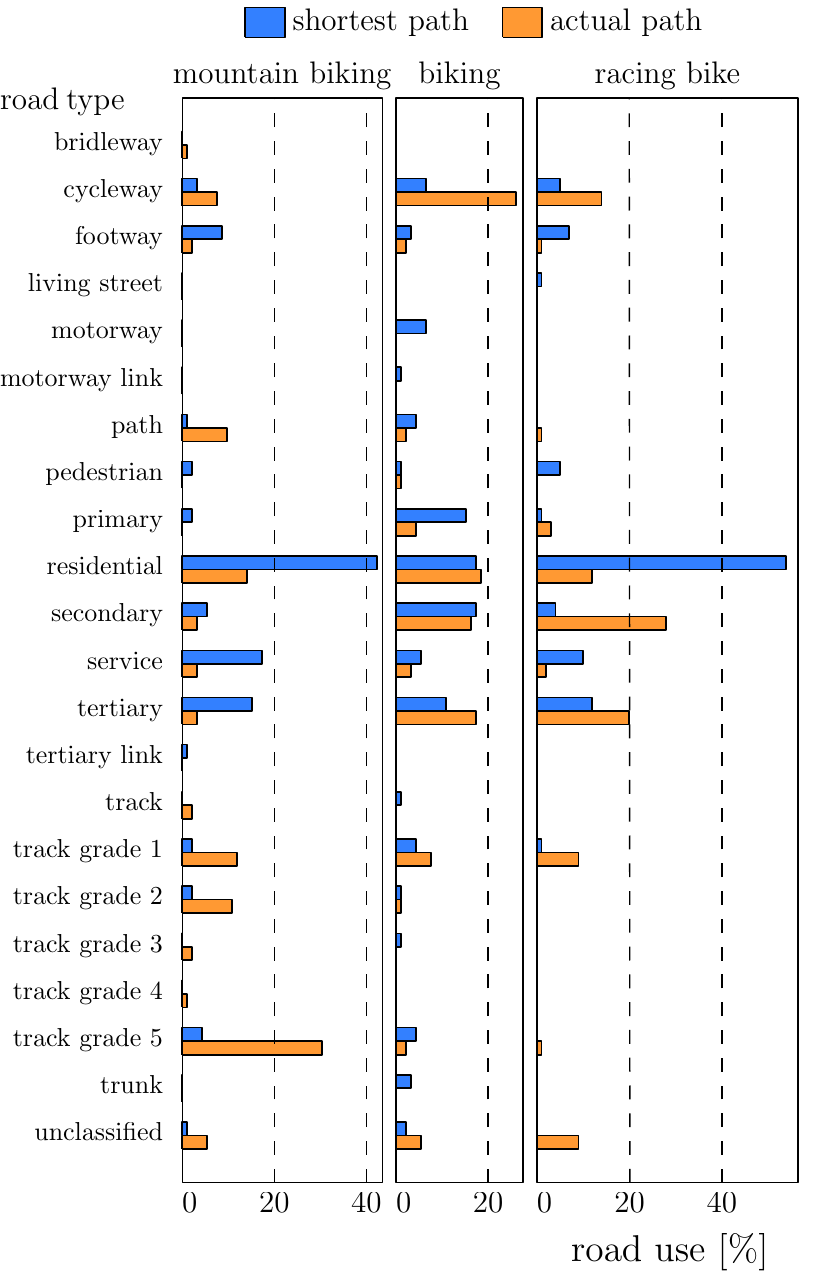}	
\caption{Share of different road types among the total length of the trajectories (orange) and the shortest paths connecting the same end nodes (blue), for each of the three types of bicyclists.}
\label{fig:road-type-classification}
\end{figure}

Figure~\ref{fig:road-type-classification} summarizes 
the share of each road type among the total length of the trajectories (i.e., the actual routes of the users) and among the total length of the shortest routes. The road type\texttt{ track grade 5}, which represents unpaved trails, 
has the largest share among the paths used by mountain bikers. 
In contrast,\texttt{ secondary }is the road type with the largest share among the paths of racing bikers.
For other cyclists,\texttt{ cycleway }is the road type with the highest usage.
These observations can be inferred from the large sizes of the corresponding orange bars in Fig.~\ref{fig:road-type-classification}.

To understand the importance of the blue bars in Fig.~\ref{fig:road-type-classification}, which represent the share of a road type among the shortest paths,
let us discuss the usage of road type\texttt{ residential }by mountain bikers.
The corresponding orange bar is relatively large (actually it comes second after the bar for\texttt{ track grade 5}) which indicates that mountain bikers quite often use residential streets.
However, the corresponding blue bar is much larger than the orange one, which means that if 
mountain bikers would plan their routes simply based on the routes' geometric lengths, they 
would end up with an extremely high usage of residential streets.
Therefore, we argue that it is legitimate to say that mountain bikers disfavor residential streets.
Similarly, based on Fig.~\ref{fig:road-type-classification}, the following conclusions are most obvious:
\begin{compactitem}
\item All groups of bicyclists disfavor footways and service streets.
\item All groups of bicyclists favor cycleways and streets of type\texttt{ track grade 1}.
\item Mountain bikers additionally prefer paths as well as the types\texttt{ track grade 2 }to\texttt{ 5},
but they disfavor residential streets and tertiary streets.
\item Racing bikers favor secondary as well as tertiary streets but disfavor residential streets.
\item Other cyclists favor tertiary streets but disfavor primary streets.
\end{compactitem}
We note that statistical tests of significance would be necessary to make more 
profound statements concerning preferred road types.
However, to obtain
a binary classification of the road types for the subsequent steps of our analysis,
it is most reasonable to apply Equations~(\ref{eq:favored}) and (\ref{eq:unfavored}).
This means, for example, that we say that users of the group\texttt{ biking }favor 
residential streets even though the share of residential streets among their routes 
is only slightly larger than among the corresponding shortest paths 
(i.e., the blue bar and the orange bar have almost the same size).

\subsection{Results of the Weight Inference}

As a final step, we applied the algorithm of \cite{OehrleinNH17} to infer a weighting for each user group. As a result we receive
for every given trajectory the size of the decomposition for every
$\alpha \in [0,1]$, in particular the size of a minimal
decomposition.

\begin{figure}[t]
  \centering
  \includegraphics[width=\columnwidth]{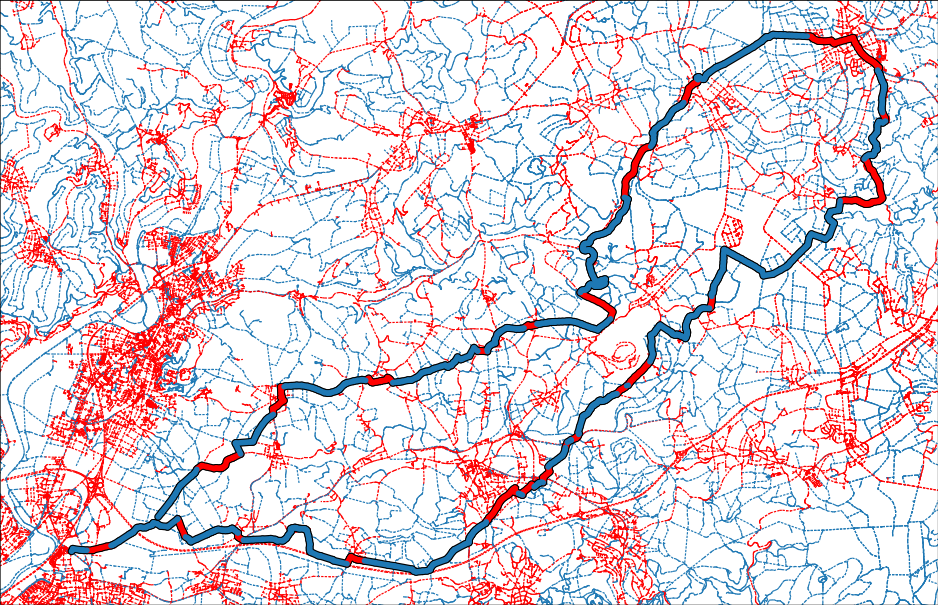}
  \caption{\label{fig:290_trajectory}A mountain bike trajectory close to Bonn. Road segments of
    favored types are depicted as blue lines, those of unfavored types
    as red lines. The trajectory (bold) has a total length of
    $51\,\mathrm{km}$ and $71.23\%$ of the trajectory is found on
    roads of favored types. }
\end{figure}

\begin{figure}[t]
  \centering
  \includegraphics[width=\columnwidth]{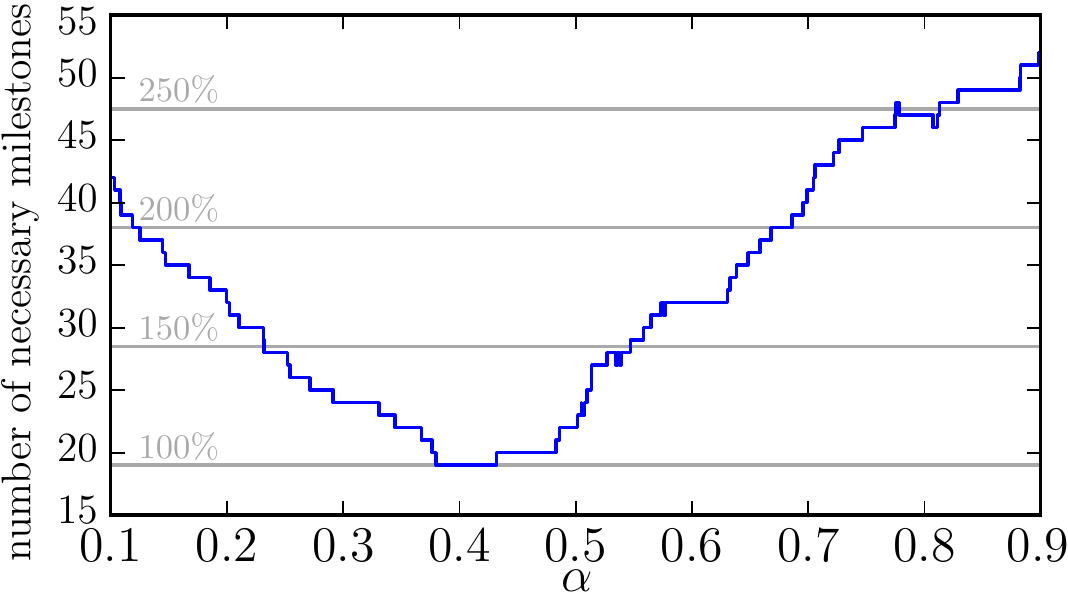}
  \caption{Analysis of the milestone decompositions of the trajectory
    given in Fig.~\ref{fig:290_trajectory}.}
  \label{fig:290_plot}
\end{figure}

\begin{figure*}
  \centering
  \includegraphics[width=\textwidth]{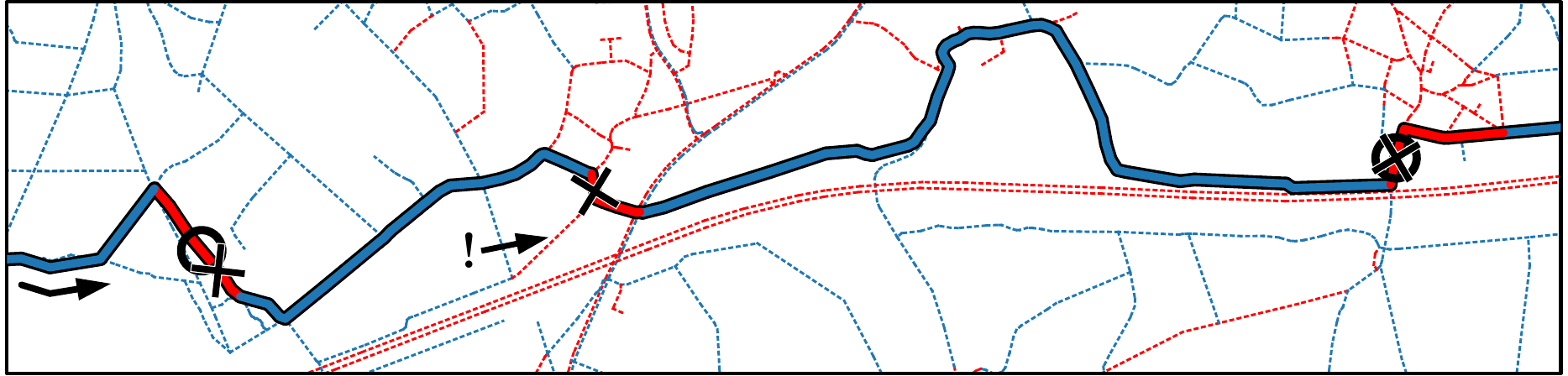}
  \caption{\label{fig:290_detail}Excerpt of the trajectory of
    Fig.~\ref{fig:290_trajectory}. Road segments of favored type are
    colored blue, those of unfavored type are colored red. For this
    subpath, a decomposition with $\alpha = 0.5$ requires three
    milestones ($\bm\times$) while already two milestones are sufficient
    with $\alpha = 0.38$ ($\bm\circ$). The road segment marked with ``!''
    causes an extra milestone for every decomposition with
    $\alpha \geq 0.48$.  Note that this implies that the subpath
    between the two circles is an optimal path for $\alpha = 0.38$ but
    not for $\alpha = 0.5$. Therefore, the edge weighting defined with
    Equation~(\ref{eq:weight}) and $\alpha = 0.38$ reflects the user's
    routing preferences relatively well (see Fig.~\ref{fig:290_plot}).}
\end{figure*}

Before analyzing the overall outcome of this step, we would like to
take a closer look at the result for a single trajectory (see
Fig.~\ref{fig:290_trajectory}). This is a nice example for a mountain
bike trajectory in a rather densely populated area: In general, the
bicyclist avoided villages and rode through the
countryside. Accordingly, the results in Fig.~\ref{fig:290_plot}
approve our classification. The number of milestones that are needed
for the decomposition is minimal for $\alpha \in [0.38,0.43]$. Such a
value for $\alpha$ means that this bicyclist accepted detours which are up
to $63\%$ longer than the shortest path in order to use favored road
types instead of unfavored ones. An example explaining this
interpretation in detail can be found in Fig.~\ref{fig:290_detail}.

\begin{figure}
  \centering
  \includegraphics[width=\columnwidth]{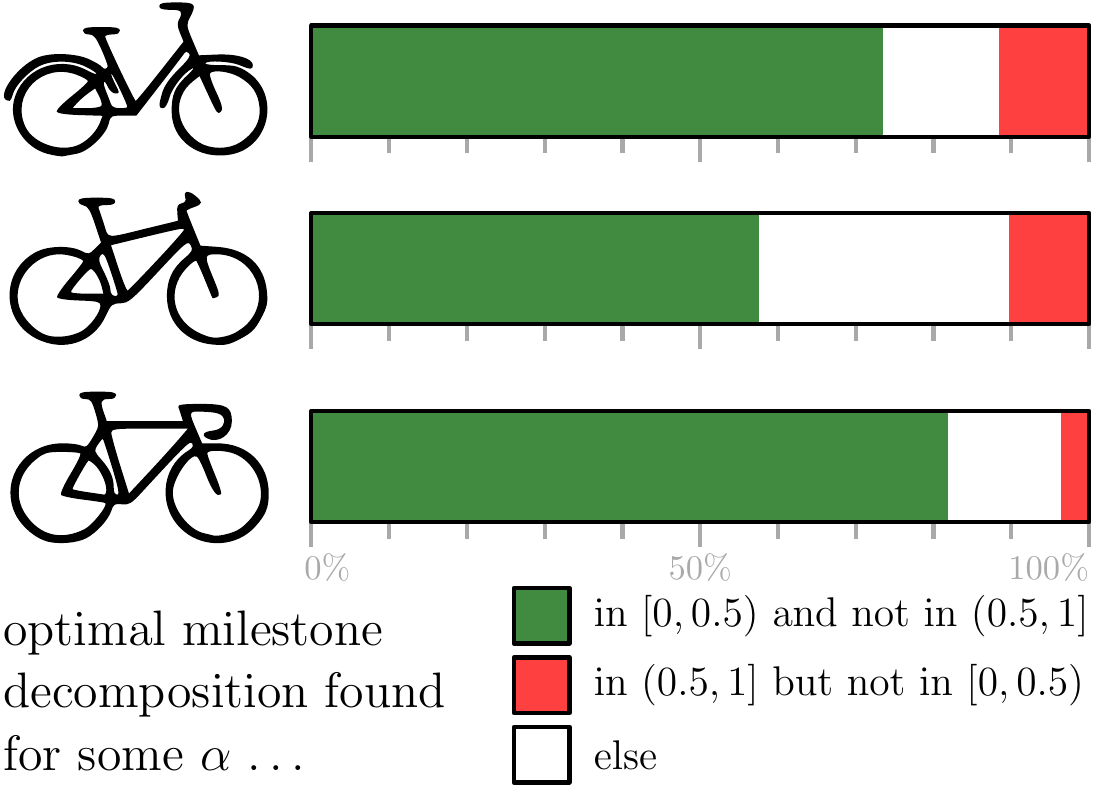}
  \caption{\label{fig:alpha_distribution} Overview of the distribution
    of minimal decompositions for \texttt{cycling} (top),
    \texttt{mountainbiking} (center), and \texttt{racingbiking}
    (bottom). The green bar indicates the share of trajectories that
    have minimal decompositions only for $\alpha$ values less than or
    equal to $0.5$; the red bar represents the trajectories with
    minimal decompositions only for $\alpha$ values greater than or
    equal to $0.5$. Please note that trajectories with an optimal
    decomposition only for $\alpha = 0.5$ as well as all remaining
    trajectories are represented by the white bar.}
\end{figure}

Thus, any trajectory that has an optimal decomposition for
$\alpha < 0.5$ approves our
classification. Fig.~\ref{fig:alpha_distribution} gives an insight how
applicable our classification is. In particular, for the user group
\texttt{racingbiking} four out of five trajectories have optimal
decompositions for $\alpha < 0.5$ but not for $\alpha > 0.5$. The weakest
classification is the one for the group \texttt{mountainbiking}. But
even here almost $60\%$ of the trajectories have a minimal decomposition certifying
our findings. This group also has the highest proportion of
trajectories that have a minimum decomposition for $\alpha > 0.5$ but not
for $\alpha < 0.5$ (roughly $10\%$).

\begin{figure}
  \includegraphics[width=\columnwidth]{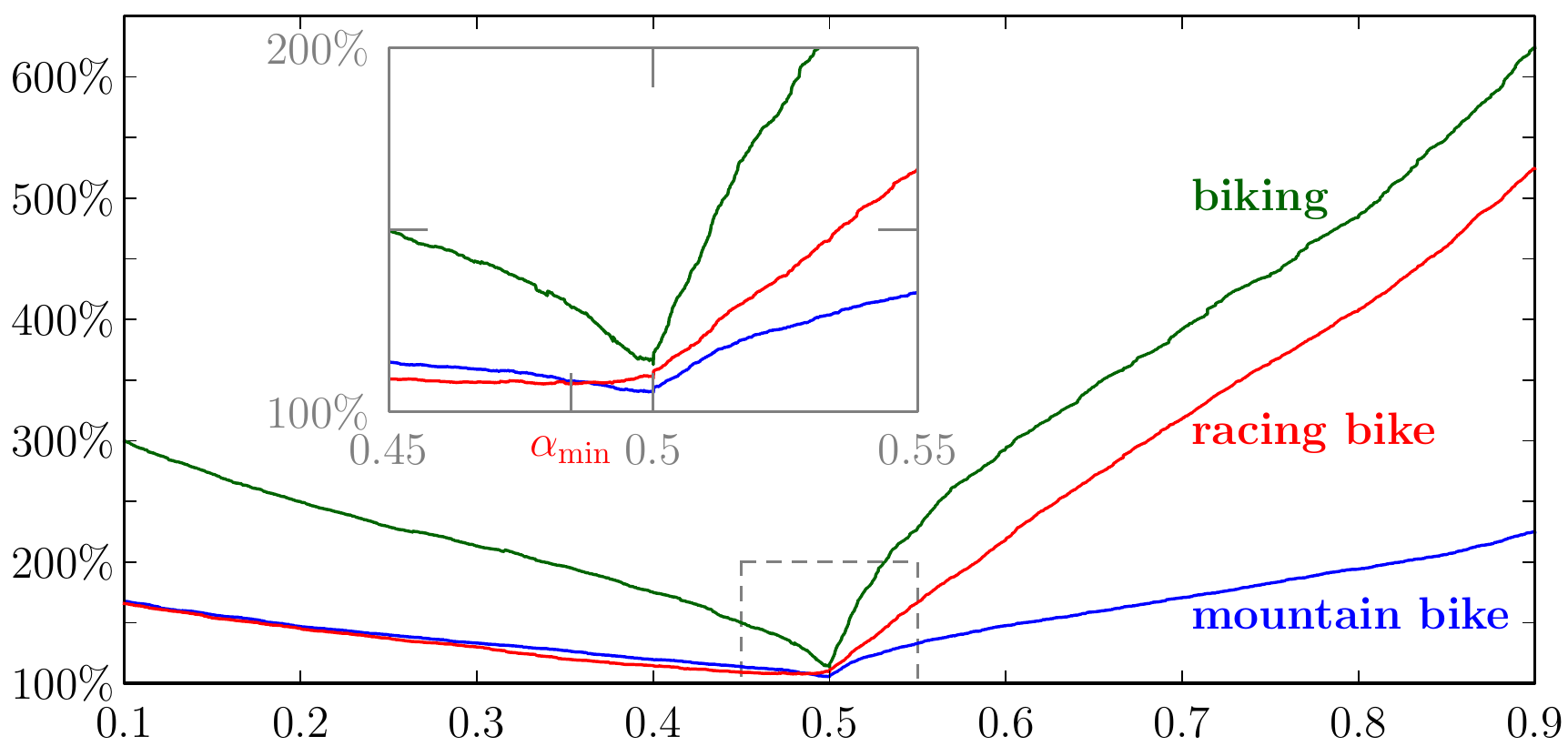}	
  \caption{\label{fig:overview_md} Number of milestones as a function
    of $\alpha$, summed over all trajectories of the same type and
    measured in percent relative to the minimum number of
    milestones. The minimum is attained close to $\alpha = 0.485$ for
    racing bike and at $\alpha = 0.5$ for biking and mountain biking.
  }
\end{figure}

For further analysis, we take the size of a minimal decomposition of a
trajectory as $100\%$ and consider for every alpha the necessary
number of milestones relative to the size of a minimal decomposition
in percent, see the gray lines in Fig.~\ref{fig:290_plot}. Finally, we compute the
average percentage of necessary milestones per $\alpha$ for every user
group. Figure~\ref{fig:overview_md} gives an overview of these
numbers. At first glance, the results are in accord with the results
of Fig.~\ref{fig:alpha_distribution} and approve our
classification. On average, focusing on favored road types is more
convenient for every user group than focusing on unfavored road
types. Even for the lowest curve, referring to the user group of
mountain bikers, it takes more than $50\%$ of milestones extra for
$\alpha = 0.9$ in comparison to $\alpha = 0.1$. Taking a closer look,
one notices that, for the group \texttt{racingbiking}, the best
results are obtained for $\alpha \approx 0.485$.  That means, that
racing bikers are willing to make detours of more than $6\%$ in order
to use road types that we have recognized as favored ones. But, for
\texttt{biking} and \texttt{mountainbiking} the number of necessary
milestones is, on average, minimal for $\alpha = 0.5$. That means,
despite an (in parts clear) classification into favored and unfavored
road types, the routing results that are best for all users within one of the two groups are achieved when ignoring the classification and simply considering distance. 
In other words, there is no value for $\alpha$ other than $0.5$
that would be better for the whole group -- this suggests 
that one should probably focus on training the parameter $\alpha$ for smaller groups 
or even for individual users.

\section{CONCLUSION}\label{sec:conclusion}
We have presented a novel approach for the classification of bicycle trajectories from crowd-sourced data into different groups. For each group (e.g., mountain bikers) we have identified favored and unfavored road types. Based on this information, we have defined a bicriteria routing model,
which assumes that bicyclists favor short routes but on the other hand try to avoid
unfavorable road types.
We have shown how the trade-off parameter of the model can be learned from the trajectories
such that a single edge weighting is obtained that can be used to compute new routes
for any two nodes in the cycle network.
To this aim, a multi-source data analysis consisting of three steps has been performed. 

In the first step, a map-matching approach has been applied in order to combine data from different sources and to extract a significant feature set for the classification of different cyclist groups in an unsupervised manner. We are discriminating between three user groups: \texttt{mountainbiking}, \texttt{racingbiking} and \texttt{biking}. Our results confirmed the user-specified groups with a consensus in 67\% of all trajectories. A feature importance analysis revealed that parameters such as the route type (such as circular or simple track), the altitude range, and the difference in length between the trajectory and the respective shortest path turn out to be of great interest for a group categorization. 

In the second step, we have identified favored and unfavored road types with regard to each of the three groups.
While some types such as \texttt{cycleway} are preferred by all groups of cyclists, the analysis also revealed  large differences among the different groups.
For example, streets of type \texttt{tertiary} are clearly favored by the groups \texttt{racingbiking} and
\texttt{biking} but clearly disfavored by the group \texttt{mountainbiking}.

In the third step, despite the sparseness of the underlying trajectory sets, we were able to learn a mapping of edge types and edge lengths to edge weights. 
The results we obtained prove that our approach goes in the right direction. 
Basically, our classification is proper but needs additional fine-tuning in order to outweigh bicyclists' demand for shortest paths.
Particularly for the group \texttt{racingbike} we succeeded and identified a mapping to edge weights that results in paths that are optimal although being $6\%$ longer than shortest paths.
For the groups \texttt{mountainbiking} and \texttt{biking} it turned out that,
if the aim is to satisfy all users equally well, the best solution to the routing problem
would be simply to minimize the geometric length of the path.
Therefore, as a direction for future research, 
we suggest considering a classification of users into more than three groups or learning the trade-off parameter
of the routing model individually for each user. Clustering algorithms such as spectral clustering \citep{ng2001spectral} or mean shift algorithm \citep{comaniciu2002mean} state promising alternatives to k-means, and could facilitate an appropriate choice of the number of clusters. 
Since we have observed that the users sometimes change their routing criteria even within single trajectories 
(e.g., since a mountain biker behaves like a normal biker when riding to or back from a hilly region of interest) it may also be reasonable to ask for a partition of a given trajectory into parts that are homogeneous with respect to the routing criteria applied. 
     
\bibliography{sdsc2018_bicyclists}

\end{document}